\documentclass[runningheads]{llncs}
\usepackage{graphicx}
\usepackage{cite}
\usepackage{amsmath,amssymb,amsfonts}
\usepackage{algorithmic}
\usepackage{svg}
\usepackage{textcomp}
\usepackage{adjustbox}
\usepackage{multirow}
\usepackage{bbold}
\usepackage{siunitx}
\usepackage{hyperref}       
\usepackage{xcolor}
\usepackage{sidecap}
\usepackage{subcaption} 
\usepackage{booktabs}
\usepackage[mathscr]{eucal}

\hypersetup{
    colorlinks,
    linkcolor={blue!50!black},
    citecolor={blue!50!black},
    urlcolor={blue!50!black}
}
\usepackage{amsmath, bm}
\usepackage{hyperref}

\usepackage{pifont}

\usepackage{floatrow}
\newfloatcommand{capbtabbox}{table}[][\FBwidth]

\begin{document}

\title{EndoSfM3D: Learning to 3D Reconstruct Any Endoscopic Surgery Scene using Self-supervised Foundation Model}
\author{Changhao Zhang\inst{1} \and
Matthew J. Clarkson\inst{1} \and
Mobarak I. Hoque\inst{1,2}}

\authorrunning{Zhang et al.}

\institute{UCL Hawkes Institute and Department of Medical Physics and Biomedical Engineering, University College London, UK \and
Division of Informatics, Imaging and Data Sciences, The University of Manchester, Manchester, UK\\
\email{changhao.zhang.24@ucl.ac.uk, m.clarkson@ucl.ac.uk, mobarak.hoque@manchester.ac.uk}}

\maketitle              
\begin{abstract}

3D reconstruction of endoscopic surgery scenes plays a vital role in enhancing scene perception, enabling AR visualization, and supporting context-aware decision-making in image-guided surgery. A critical yet challenging step in this process is the accurate estimation of the endoscope's intrinsic parameters. In real surgical settings, intrinsic calibration is hindered by sterility constraints and the use of specialized endoscopes with continuous zoom and telescope rotation. Most existing methods for endoscopic 3D reconstruction do not estimate intrinsic parameters, limiting their effectiveness for accurate and reliable reconstruction. In this paper, we integrate intrinsic parameter estimation into a self-supervised monocular depth estimation framework by adapting the Depth Anything V2 (DA2) model for joint depth, pose, and intrinsics prediction. We introduce an attention-based pose network and a Weight-Decomposed Low-Rank Adaptation (DoRA) strategy for efficient fine-tuning of DA2. Our method is validated on the SCARED and C3VD public datasets, demonstrating superior performance compared to recent state-of-the-art approaches in self-supervised monocular depth estimation and 3D reconstruction. Code and model weights can be found in project repository: \href{https://github.com/MOYF-beta/EndoSfM3D}{https://github.com/MOYF-beta/EndoSfM3D}


\end{abstract}

\section{Introduction}

Depth estimation plays a crucial role in minimally invasive endoscopic surgery by supporting precise navigation, accurate 3D surface reconstruction, and realistic augmented-reality visualization~\cite{collins2020augmented, zhang2020real}. Achieving reliable depth estimation in surgical scenes, however, is difficult due to complex anatomical structures, limited illumination, and texture-sparse regions~\cite{shao2022self}. Classical multi-view geometry techniques such as Structure from Motion (SfM)~\cite{rattanalappaiboon2015fuzzy} and Simultaneous Localization and Mapping (SLAM)~\cite{grasa2013visual} often fail under such conditions. Deep learning approaches have shown strong performance for depth estimation in natural images~\cite{sun2023sc, bhat2022localbins}, but obtaining large-scale, high-quality surgical depth ground truth for supervised learning is constrained by data privacy, security, and clinical expertise requirements. Consequently, self-supervised learning (SSL) approaches have gained attention, where depth predictions are guided by geometric consistency across video frames~\cite{arampatzakis2023monocular, ozyoruk2021endoslam}. Shao \textit{et al.}~\cite{shao2022self} addressed illumination inconsistencies in endoscopic depth estimation using appearance flow. Yang \textit{et al.}~\cite{yang2024self} proposed a compact architecture integrating CNN and Transformer modules to reduce model complexity. Zeinoddin \textit{et al.}~\cite{zeinoddin2024dares} employed the pre-trained Depth Anything V2 model with Parameter-Efficient Fine-Tuning (PEFT) for surgical depth estimation, while Cui \textit{et al.}~\cite{cui2024endodac} adopted a pre-trained DinoV2 encoder with a novel Dynamic Vector-Based Low-Rank Adaptation (DV-LoRA) strategy. However, in the aforementioned studies, researchers have predominantly focused on the accuracy of monocular depth estimation, often overlooking a critical factor for its application to 3D reconstruction.

Unlike conventional imaging environments, endoscopic camera calibration remains particularly challenging due to dynamically changing imaging parameters during surgery. Optical zooming causes continuous variation in intrinsic parameters such as focal length, principal point, and distortion coefficients, making traditional calibration methods unreliable \cite{an2021two}. In addition, telescope rotation in oblique-viewing endoscopes alters projection geometry and extrinsic pose relationships, introducing further calibration instability \cite{eppenga2023}. Manual endoscope manipulation compounds these issues through hand tremor–induced errors and inconsistent positioning \cite{alobaidia2017}. These challenges are further intensified by surgical constraints such as limited field of view, occlusions, and sterility requirements \cite{mountney2011}. Despite advances in modeling and calibration, achieving robust and accurate estimation under continuous zoom and rotation remains an open problem with substantial clinical relevance. In this work, we observe that the re-projection loss used in SfM-based self-supervised learning can be extended to jointly optimize the intrinsic parameter matrix, offering an alternative route for precise intrinsic estimation.

Our key contributions can be summarized as follows: 
(1)~We integrate intrinsic parameter estimation into the AF-SfM self-supervised monocular depth estimation framework by adapting the Depth Anything V2 (DA2) model for \textit{joint prediction} of depth, pose, and intrinsics, addressing a critical limitation in endoscopic 3D reconstruction where sterility constraints and specialized endoscopes hinder traditional calibration. 
(2)~We introduce two key architectural innovations: an attention-based pose network and a Weight-Decomposed Low-Rank Adaptation (DoRA) \cite{liu2024dora} strategy for efficient fine-tuning of DA2, optimizing parameter efficiency while maintaining representation capacity. 
(3)~Validation on the SCARED and C3VD datasets demonstrates state-of-the-art performance, achieving best-reported AbsRel scores of \textbf{0.050} (SCARED) and \textbf{0.058} (C3VD) for depth estimation, outperforming recent methods in self-supervised endoscopic reconstruction on depth estimation accuracy, and accurate intrinsic prediction with error of less than \textbf{2\%} on critical intrinsic parameter focal length $( f_x,f_y )$ and less than 10\% on principal point coordinate $( c_x,c_y )$.

\section{Method}

\begin{figure}[ht]
\centering
\includegraphics[width=1\linewidth]{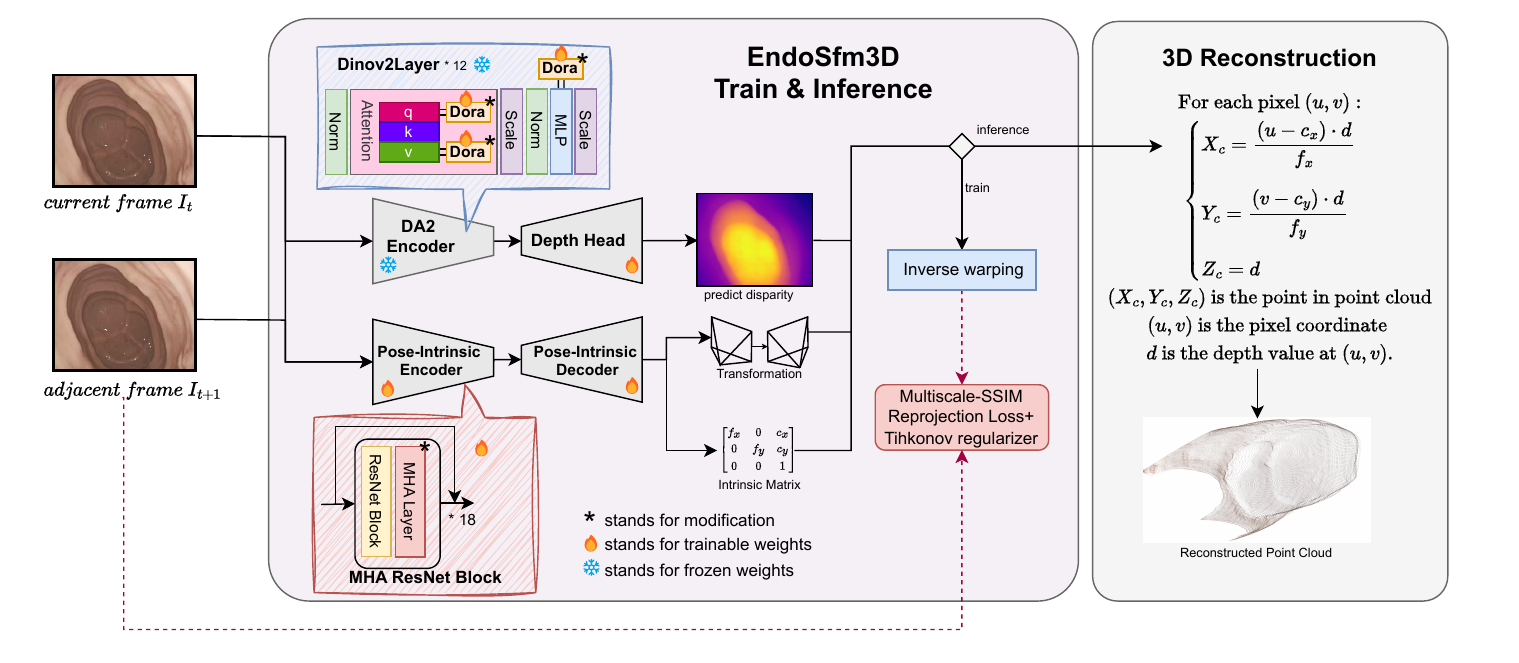}
\caption{Illustration of the proposed EndoSfM3D framework. A ViT-based encoder and DPT-like decoder pre-trained from Depth Anything V2 \cite{yang2024depth} are used for depth estimation. Weight-Decomposed Low-Rank Adaptation (DoRA) is applied to the feature extractor for parameter-efficient fine-tuning (PEFT). A ResNet-based Pose Encoder with multi-head attention layers extracts SfM features shared between two frames for transformation and intrinsic estimation. Both depth and pose networks are trained using the reprojection loss from MonoDepth2 \cite{godard2019digging} and the Tikhonov regularizer from AF-SfMLearner \cite{shao2022self}.}
\label{fig:arch}
\end{figure}

\subsection{Preliminaries}

\subsubsection{Weight-Decomposed Low-Rank Adaptation (DoRA)~\cite{liu2024dora}} 
DoRA was proposed to enhance the fine-tuning efficiency of foundation models by decoupling the weight matrix into directional and magnitude components. It is motivated by the intuition that directional updates capture task-specific feature orientations while magnitude adjustments regulate their intensities, enabling more precise adaptation. DoRA achieves parameter-efficient tuning by decomposing each pre-trained weight matrix into a directional matrix and a trainable magnitude vector, combined with low-rank updates. Specifically, for a pre-trained weight matrix $W_{0}\in \mathbb{R}^{d \times k}$, DoRA reformulates the computation as:
\begin{equation}
h = \left(\frac{V}{||V||} \odot g\right)x + BAx = \left(\frac{W_0 + \Delta V}{||W_0 + \Delta V||} \odot g\right)x + BAx
\end{equation}
where $V = W_0 + \Delta V$ represents the directional component with $||\cdot||$ denoting L2 normalization, $g\in \mathbb{R}^{d}$ is a learnable magnitude vector, and $B\in \mathbb{R}^{d \times r}, A\in \mathbb{R}^{r \times k}$ are low-rank matrices ($r \ll \min(d,k)$). Only $\Delta V$, $g$, $A$, and $B$ are updated during training, while the original weights $W_0$ remain frozen.

\subsubsection{Self-supervised Depth, Ego-motion, and Intrinsic Estimation}
Self-supervised methods for depth and ego-motion estimation leverage view synthesis as a supervisory signal. Given consecutive frames $I_t$ (target) and $I_s$ (source), the core objective is to minimize the photometric error between $I_t$ and the warped source image $I_{s \to t}$. The warping operation $\pi$ uses estimated depth $z$, camera intrinsics $K$, and relative pose $(R,t)$ as defined in Equation~\ref{equation:reproj}:

\begin{equation}
I_{s\to t} = \pi \left( z,K,R,t,I_{s}\right)
\end{equation}

The photometric loss combines L1 and SSIM metrics:
\begin{equation}
\mathcal{L}_p = \alpha \frac{1-\operatorname{SSIM}\left(I_t, I_{s \to t}\right)}{2} + (1-\alpha)\left\|I_t - I_{s \to t}\right\|_1
\label{eq:photometric}
\end{equation}

\subsubsection{Tikhonov Regularizer}

To stabilize self-supervised depth and ego-motion learning in endoscopy, we use a Tikhonov regularizer $\mathcal{R}_k$ following~\cite{shao2022self}. It incorporates smoothness and consistency priors, helping to address the problem's ill-posed nature. The regularizer is defined as:
\begin{equation}
\mathcal{R}_k = \lambda_1 \mathcal{L}_{rs} + \lambda_2 \mathcal{L}_{ax}
\label{eq:kiv}
\end{equation}

Compared with Af-SfMLearner's Tikhonov regularizer\cite{shao2022self}, we removed Edge-Aware Smoothness Loss $ \mathcal{L}_{es}$, as it's redundant and harmful to powerful pre-trained depth model We describe the first two components below:

\paragraph{(a) Residual-Based Smoothness Loss $\mathcal{L}_{rs}$}
This term enforces smoothness in the predicted appearance flow $\mathbf{C}_\delta$, while preserving edges near high photometric differences (e.g., specular regions). It is defined as, where $\mathbf{p} = (u,v)$ is the coordinate of pixel:
\begin{equation}
\mathcal{L}_{rs} = \sum_{\mathbf{p}} |\nabla \mathbf{C}_\delta(\mathbf{p})| \cdot e^{-\nabla |I^t(\mathbf{p}) - I^{s \to t}(\mathbf{p})|}
\end{equation}
The exponential term reduces smoothness penalties in areas with strong brightness changes.

\paragraph{(b) Auxiliary Loss $\mathcal{L}_{ax}$}
This loss aligns the appearance flow with optical flow to ensure consistency with motion cues:
\begin{equation}
\mathcal{L}_{ax} = \sum_{\mathbf{p}} \mathbf{V}(\mathbf{p}) \cdot \Phi\left(I^{s \to t}(\mathbf{p}), I^t(\mathbf{p}) + \mathbf{C}_\delta(\mathbf{p})\right)
\end{equation}
Here, $\Phi(\cdot)$ measures photometric error, and $\mathbf{V}(\mathbf{p})$ masks out occluded regions. This encourages $\mathbf{C}_\delta$ to follow realistic motion patterns.

\subsection{Proposed Framework: EndoSfM3D}
As illustrated in Fig.~\ref{fig:arch}, The EndoSfM3D fine-tune the Depth Anything V2 model into an Endoscope foundation model for accurate depth estimation. The framework includes three main components: DepthNet, Pose-Intrinsics Net, and a semi-supervised training system based on EndoSfM3D we proposed. DepthNet predicts depth from the reference frame using a ViT-based encoder (pre-trained with Depth Anything V2) and a DPT-like decoder. Pose-Intrinsics Net predicts motion and camera parameters from the same input, sharing a ResNet encoder with two output heads. EndoSfM3D uses predicted depth, motion, and camera parameters to reproject frames and optimize the model by comparing reprojected and reference frames, thus getting the self-supervised training loss.  In EndoSfM3D,  compared with original AF-SfMLearner~\cite{shao2022self} which introduced multiple regularizer to enhance the depth estimation, we only kept smoothness of optic flow and disparity, removed other constrains since a powerful pretrained depth model no longer need them. In optimizer, we utilize $Loss = \mathcal{L}_p + \mathcal{R}_k$ where  $\mathcal{L}_p$ and $\mathcal{R}_k$ are defined by we utilize $Loss = \mathcal{L}_p + \mathcal{R}_k$ where  $\mathcal{L}_p$ and $\mathcal{R}_k$ are defined by (\ref{eq:photometric}) and (\ref{eq:kiv}).

\subsubsection{Pretrained Foundation Models for Depth} Foundation Models generally refer to powerful pre-trained models trained on extensive amounts of data which enable them to exhibit strong generalization capabilities across multiple tasks and scenarios. Dense Prediction Transformer (DPT)~\cite{ranftl2021vision} is a depth estimation foundation model based on Vision Transformer (ViTs). DINOv2~\cite{oquab2023dinov2} is a semantic foundation model suitable for many vision tasks including depth estimation with separate decode decoders. In this work, we aim to adapt Depth Anything~\cite{yang2024depth}, which is a depth estimation foundation model trained on large-scale labeled and unlabeled data, to endoscopic scenes.

\subsubsection{Pose-Intrinsics Net}

The architecture employs a ResNet-18 backbone, optimized for computational efficiency and feature extraction in endoscopic applications, processing pairs of consecutive frames for pose estimation. Multi-Head Attention (MHA) modules are integrated after each ResNet block (layer1–layer4), enabling spatial relationship learning across feature resolutions of 64, 128, 256, and 512 channels. Each MHA module uses 8 attention heads, where Query, Key, and Value matrices are generated via 1×1 convolutions on ResNet outputs. Each head processes $channel_{dim}/8$ channels, focusing attention on spatially relevant regions.

Feature extraction occurs in two stages: standard ResNet blocks initially process features, followed by MHA modules with residual connections. These modules apply layer normalization and self-attention, with outputs added to input features via residuals. This design simultaneously captures spatial attention patterns critical for both ego-motion and intrinsic parameter prediction. The component takes consecutive frames as input, predicting 6DoF ego-motion and camera intrinsics through separate heads. These predictions—combined with depth estimates—enable pixel-wise 3D re-projection through geometric reprojection. The re-projection formula is as follows:

\begin{equation}
z^{'} p^{'} = KRK^{-1}zp + Kt, 
\label{equation:reproj}
\end{equation}
where K refers to the intrinsic matrix given by 
$
K = \begin{vmatrix}
f_{x} & 0 & x_{0} \\
0 & f_{y} & y_{0} \\
0 & 0 & 1 \\
\end{vmatrix},
\label{equation:K}
$
$p$ and $p^{'}$ are pixel coordinates before and after the transformation of rotation matrix $R$ and translation vector $t$; $z$ and $z^{'}$ are corresponding depths. Previous work~\cite{gordon2019depth} has demonstrated that given Equation.~(\ref{equation:reproj}), no $\tilde{K}$ and $\tilde{R}$ exist such that $\tilde{K}\tilde{R}\tilde{K}^{-1} = KRK^{-1}$ leading the estimation of $K$, $R$ and $t$ to converge simultaneously. Therefore, the ego-motion head and the intrinsic parameters head share the same encoder.

\subsection{Experiment and Results}

\subsubsection{Dataset and Implementation Details}
We trained and validated our model on two datasets: SCARED~\cite{allan2021stereo} and C3VD~\cite{bobrow2023colonoscopy}. 
SCARED contains 35 porcine abdominal endoscopy videos (22,950 frames) with depth, pose, and intrinsic ground truth, using Shao et al.'s split \cite{shao2022self}. 
C3VD provides 22 synthetic colon model sequences (10,015 frames) with comprehensive ground truth annotations, following Paruchuri et al.'s split \cite{paruchuri2024leveraging}.
During training, we employ the Adam optimizer with an initial learning rate of $1\times10^{-4}$, which decays to $1\times10^{-5}$ after 10 epochs, and train all models for 20 epochs with batch size 12. The training loss is given by $Loss = \mathcal{L}_p + \mathcal{R}_k$ where  $\mathcal{L}_p$ and $\mathcal{R}_k$ are defined by we utilize $Loss = \mathcal{L}_p + \mathcal{R}_k$ where  $\mathcal{L}_p$ and $\mathcal{R}_k$ are defined by (\ref{eq:photometric}) and (\ref{eq:kiv}) respectively. respectively.

\begin{table}[ht]
\begin{floatrow}
\capbtabbox{
\resizebox{0.7\textwidth}{!}{
\begin{tabular}{c|c|c|cccc}
                    \toprule  
                     & \textbf{Method} & \textbf{Year} & \textbf{AbsRel$\downarrow$} & \textbf{SqRel$\downarrow$} & \textbf{Rmse$\downarrow$} & \textbf{RmseLog$\downarrow$}\\ 
                    \midrule
                    \multirow{7}{*}{\rotatebox{90}{SCARED}} 
                    & Fang et al.~\cite{fang2020towards} & 2020 & 0.078 & 0.794 & 6.794 & 0.109\\
                    & Monodepth2~\cite{godard2019digging} & 2019 & 0.069 & 0.577 & 5.546 & 0.094 \\
                    & Endo-SfM~\cite{ozyoruk2021endoslam} & 2021 & 0.062 & 0.606 & 5.726 & 0.093 \\
                    & AF-SfMLearner~\cite{shao2022self} & 2022 & 0.059 & 0.435 & 4.925 & 0.082 \\ 
                    & DARES\cite{zeinoddin2024dares} & 2024 & 0.052 & 0.356 & 4.483 & 0.073 \\
                    & EndoDAC\cite{cui2024endodac} & 2024 & 0.052 & 0.362 & \textbf{4.464} & 0.073 \\
                    & Endo-FASt3r\cite{zeinoddin2025endo} & 2025 & 0.051 & \textbf{0.354} & 4.480 & - \\
                    & \textbf{EndoSfM3D (Ours)} & - & \textbf{0.050} & 0.389 & 4.749 & \textbf{0.070} \\ 
                    \midrule
                    \multirow{3}{*}{\rotatebox{90}{C3VD}} & AF-SfMLearner~\cite{shao2022self} & 2022 & 0.086 & 0.358 & - & 0.104 \\ 
                    & Col3D-MTL \cite{solano2025multi} & 2023 & 0.109 & 0.386 & \textbf{3.052} & 0.131 \\ 
                    & \textbf{EndoSfM3D (Ours)} & - & \textbf{0.058} & \textbf{0.333} & 4.413 & \textbf{0.076} \\ 
                    \bottomrule
                \end{tabular}}
}{
 \caption{Quantitative depth estimation comparison on SCARED and C3VD datasets. Bold reflect the best result.} 
 \label{tab:depth}
} 

\end{floatrow}
\end{table}

\begin{figure}[ht]
    \centering
    \includegraphics[width=1\linewidth]{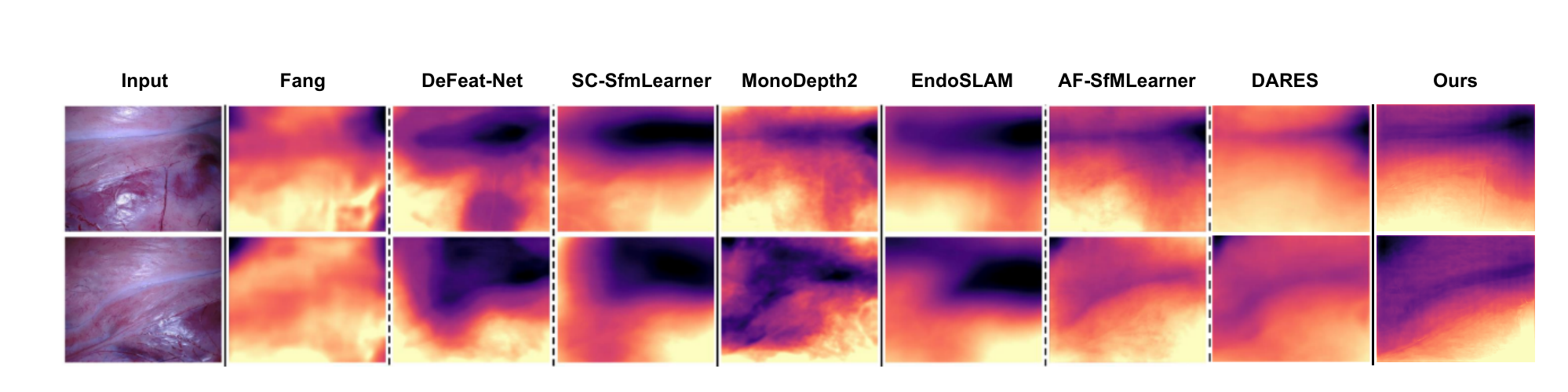}
    \caption{Results comparison on SCARED for previous researches and our results.}
    \label{fig:compare_full}
\end{figure}

\begin{figure}[ht]
\includegraphics[width=0.7\linewidth]{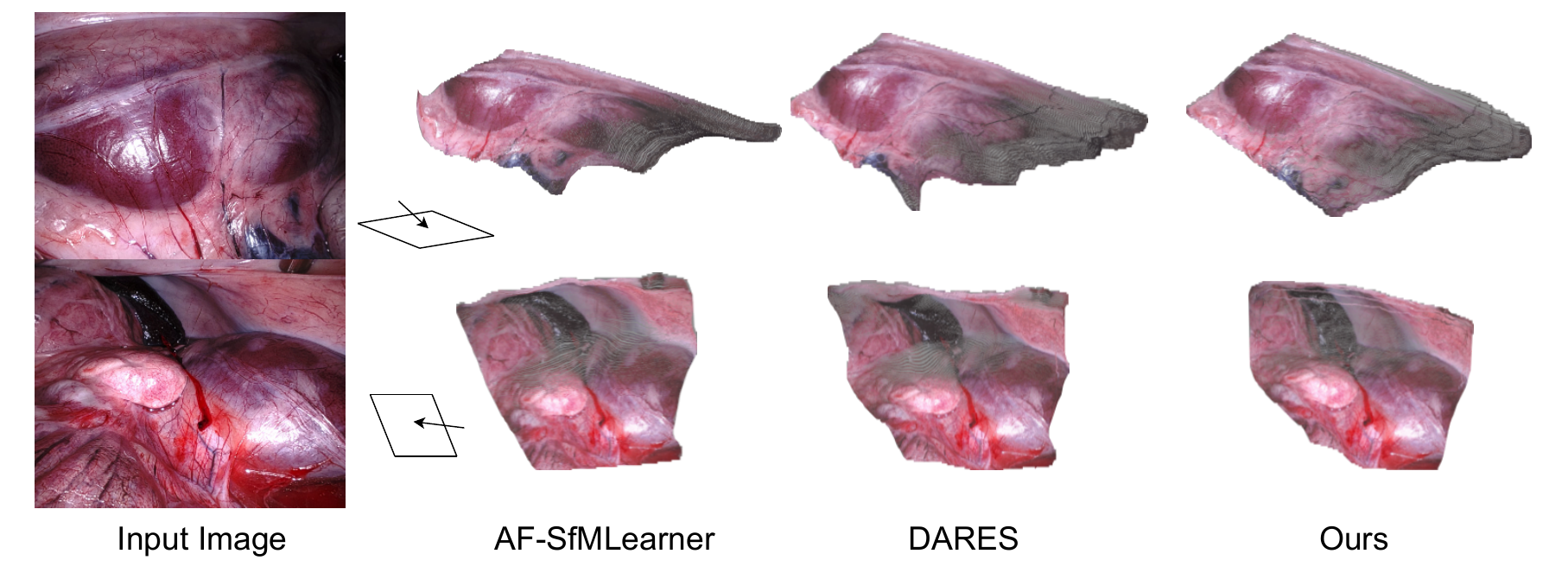}
\captionof{figure}{Comparison of 3D reconstruction results using AF-SfMLearner\cite{shao2022self}, DARES\cite{zeinoddin2024dares}, and our method.}
\label{fig:3d_sub}
\end{figure}

\noindent \textbf{Depth Estimation.} The proposed method is compared with several SOTA self-supervised and supervised methods~\cite{fang2020towards, spencer2020defeat, bian2019unsupervised, godard2019digging, ozyoruk2021endoslam, shao2022self, yang2024self, recasens2021endo,zeinoddin2024dares,cui2024endodac,solano2025multi}, and Depth Anything ~\cite{yang2024depth} model, all results corresponding to the trained dataset comes from their paper's result. During training, we utilized 2 datasets: C3VD and SCARED. The quantitative results for depth comparison is shown on Table~\ref{tab:depth}, our method have SOTA performance on both dataset(on Abs Rel).  Fig.~\ref{fig:compare_full} shows the qualitative depth comparison between our method and other methods. Fig.~\ref{fig:3d_sub} shows several 3D reconstruction qualitative results comparison on DARES dataset, we can observe the corresponding 3D reconstruction precisely reflect the structure of organs in our method, and getting more reasonable result on the edge of images. 

\begin{figure}[h]
    \centering
    \includegraphics[width=1\linewidth]{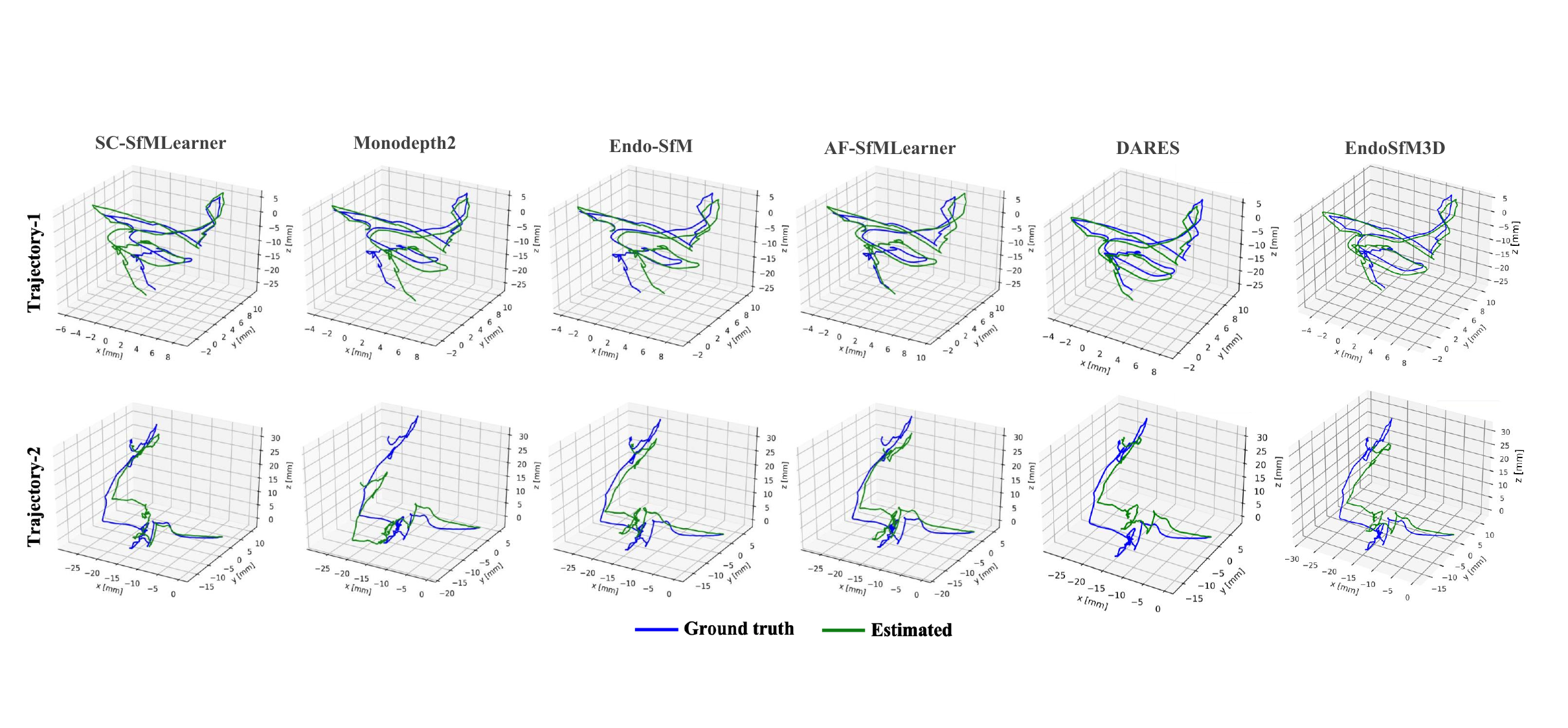}
    \caption{Trajectory estimation results on 2 SCARED sequences.}
    \label{fig:traj_sub}
\end{figure}

\noindent \textbf{Pose and Intrinsics Estimation.} 
Two sequences of SCARED dataset are selected followed ~\cite{godard2019digging, ozyoruk2021endoslam, shao2022self,zeinoddin2024dares,cui2024endodac,zeinoddin2025endo} for evaluation of pose and intrinsic estimation. Pose estimation is evaluated on 2 sequences separately while intrinsic estimation is evaluated with a weighted average percentage error on two sequences. The results are presented in Table~\ref{tab:pose} and Table~\ref{tab:intrinsics}. Table~\ref{tab:pose} shows our proposed method obtains satisfactory performances on pose estimation with or without given intrinsic parameters, with Fig.\ref{fig:traj_sub} shows the qualitative results on 2 SCARED sequence to confirm the overall accuracy of pose estimation. Our proposed method can also estimate accurate camera intrinsic parameters with error of less than \textbf{2\%} on critical intrinsic parameter focal length $( f_x,f_y)$  and less than 10\% on principal point coordinate $( c_x,c_y)$.

\begin{table}[ht]
\begin{floatrow}
\capbtabbox{
\resizebox{0.5\textwidth}{!}{
\begin{tabular}{c|c|c|c}
    \toprule  
    Method & Year & ATE $\downarrow$ (Seq.1) & ATE $\downarrow$ (Seq.2) \\ \midrule 
    Monodepth2~\cite{godard2019digging} & 2019 & 0.0769 & 0.0554 \\
    Endo-SfM~\cite{ozyoruk2021endoslam} & 2021 & 0.0759 & 0.0500 \\ 
    AF-SfMLearner~\cite{shao2022self} & 2022 & 0.0742 & 0.0478\\ 
    EndoFASt3r ~\cite{zeinoddin2025endo} & 2025 & \textbf{0.0702} & \textbf{0.0438} \\
    \textbf{ EndoSfM3D(Ours)} & - & 0.0791 & 0.0529 \\ 
    \bottomrule
\end{tabular}}
}{
 \caption{Quantitative pose estimation comparison on two selected sequences.}
 \label{tab:pose}
}

\capbtabbox{
\resizebox{0.45\textwidth}{!}{
\begin{tabular}{c|c|c|c}
\toprule
\textbf{Intrinsics} & \textbf{GT} & \textbf{Seq 1 Error} & \textbf{Seq 2 Error} \\
\midrule
$f_x$ & 769.24 & 0.68 \% & 1.56\% \\
$f_y$ & 769.24 & 0.69 \% & 1.62\% \\
$c_x$ & 679.54 & 7.73 \% & 5.42\% \\
$c_y$ & 543.97 & 10.92\% & 0.80\% \\
\bottomrule
\end{tabular}}
}{
 \caption{Intrinsic estimation comparison on two selected C3VD sequences.}
 \label{tab:intrinsics}
}
\end{floatrow}
\end{table}

\begin{table}[!ht]
\centering
\resizebox{0.7\textwidth}{!}{
\begin{tabular}{c|c|c|cccc}
\toprule  
\textbf{Depth} & \textbf{Pose} & \textbf{Intrinsic} & \textbf{AbsRel$\downarrow$} & \textbf{SqRel$\downarrow$} & \textbf{Rmse$\downarrow$} & \textbf{RmseLog$\downarrow$} \\ 
\midrule
LoRA & ResNet & Fixed& 0.0523 & \underline{0.383} & \underline{4.6635} & 0.076 \\
LoRA & ResNet & \underline{Predicted}  & 0.0527 & 0.436 & 4.9580 & 0.079 \\
LoRA & \underline{MHA ResNet} & Fixed  & 0.0509 & 0.387 & 4.6837 & 0.074 \\
\underline{DoRA} & \underline{MHA ResNet} & Fixed  & \textbf{0.0501} & \textbf{0.357} & \textbf{4.511} & \underline{0.072} \\ 
\underline{DoRA} & \underline{MHA ResNet} & \underline{Predicted}  & \underline{0.0504} & 0.389 & 4.749 & \textbf{0.070} \\
\bottomrule
\end{tabular}}
\caption{Ablation study on DARES dataset comparing depth estimation variants. Underlined methods denote framework adaptations; bold/underlined results indicate best/second-best performance.}
\label{tab:Ablation}
\end{table}

\subsubsection{Ablation study}
Table \ref{tab:Ablation} presents the ablation results. We progressively enabled the proposed components, including LoRA/DoRA integration, variations of MHA layers in the Pose-Intrinsics Net, and intrinsic parameter prediction, while monitoring performance changes. The results show consistent improvements with each modification, indicating that the proposed architectural choices contribute effectively to performance. Moreover, the inclusion of intrinsic prediction has a minimal effect on overall accuracy, confirming its stability within the framework.

\section{Conclusion}

In this work, we introduced EndoSfM3D, a self-supervised framework that unifies depth, pose, and intrinsic parameter estimation for endoscopic 3D reconstruction. The method addresses a long-standing challenge in surgical vision by enabling accurate intrinsic calibration through geometric consistency within monocular video, removing the need for physical calibration targets. Experiments on multiple endoscopic datasets show that EndoSfM3D achieves reliable depth and motion estimation while maintaining stable intrinsic predictions under zoom and rotation. This joint optimization enables accurate 3D scene reconstruction from standard surgical videos and extends the practicality of self-supervised learning to real clinical environments. Future work will focus on improving temporal consistency, domain generalization across different endoscope types, and integration with real-time image-guided surgery systems to enhance clinical applicability.

\begin{credits}
	\subsubsection{\ackname} This work was supported in whole, or in part, by the Engineering and Physical Sciences Research Council (EPSRC) under grants [EP/W00805X/1]. For the purpose of open access, the author has applied a CC BY public copyright licence to any author accepted manuscript version arising from this submission.
	
	\subsubsection{\discintname} The authors have no competing interests to declare that are relevant to the content of this article.
\end{credits}

\bibliography{Paper-0225}{}

\begin{thebibliography}{10}
\providecommand{\url}[1]{\texttt{#1}}
\providecommand{\urlprefix}{URL }
\providecommand{\doi}[1]{https://doi.org/#1}

\bibitem{alobaidia2017}
Al-Obaidi, A.e.a.: Effective calibration of an endoscope to an optical tracking system for medical augmented reality. Int. J. Comput. Assist. Radiol. Surg.  \textbf{12}(9),  1619--1628 (2017)

\bibitem{allan2021stereo}
Allan, M., Mcleod, J., Wang, C., Rosenthal, J.C., Hu, Z., Gard, N., Eisert, P., Fu, K.X., Zeffiro, T., Xia, W., et~al.: Stereo correspondence and reconstruction of endoscopic data challenge. arXiv preprint arXiv:2101.01133  (2021)

\bibitem{an2021two}
An, P., Ma, J., Ma, T., Fang, B., Yu, K., Liu, X., Zhang, J.: Two-point calibration method for a zoom camera with an approximate focal-invariant radial distortion model. Journal of the Optical Society of America A  \textbf{38}(4),  504--514 (2021)

\bibitem{arampatzakis2023monocular}
Arampatzakis, V., Pavlidis, G., Mitianoudis, N., Papamarkos, N.: Monocular depth estimation: A thorough review. IEEE Transactions on Pattern Analysis and Machine Intelligence  (2023)

\bibitem{bhat2022localbins}
Bhat, S.F., Alhashim, I., Wonka, P.: Localbins: Improving depth estimation by learning local distributions. In: European Conference on Computer Vision. pp. 480--496. Springer (2022)

\bibitem{bian2019unsupervised}
Bian, J., Li, Z., Wang, N., Zhan, H., Shen, C., Cheng, M.M., Reid, I.: Unsupervised scale-consistent depth and ego-motion learning from monocular video. Advances in neural information processing systems  \textbf{32} (2019)

\bibitem{bobrow2023colonoscopy}
Bobrow, T.L., Golhar, M., Vijayan, R., Akshintala, V.S., Garcia, J.R., Durr, N.J.: Colonoscopy 3d video dataset with paired depth from 2d-3d registration. Medical image analysis  \textbf{90},  102956 (2023)

\bibitem{collins2020augmented}
Collins, T., Pizarro, D., Gasparini, S., Bourdel, N., Chauvet, P., Canis, M., Calvet, L., Bartoli, A.: Augmented reality guided laparoscopic surgery of the uterus. IEEE Transactions on Medical Imaging  \textbf{40}(1),  371--380 (2020)

\bibitem{cui2024endodac}
Cui, B., Islam, M., Bai, L., Wang, A., Ren, H.: Endodac: Efficient adapting foundation model for self-supervised depth estimation from any endoscopic camera. In: International Conference on Medical Image Computing and Computer-Assisted Intervention. pp. 208--218. Springer (2024)

\bibitem{eppenga2023}
Eppenga, R.e.a.: An improved camera model for oblique-viewing laparoscopes: high reprojection accuracy independent of telescope rotation. Phys. Med. Biol.  \textbf{68}(18),  185007 (2023)

\bibitem{fang2020towards}
Fang, Z., Chen, X., Chen, Y., Gool, L.V.: Towards good practice for cnn-based monocular depth estimation. In: Proceedings of the IEEE Winter Conference on Applications of Computer Vision. pp. 1091--1100 (2020)

\bibitem{godard2019digging}
Godard, C., Mac~Aodha, O., Firman, M., Brostow, G.J.: Digging into self-supervised monocular depth estimation. In: Proceedings of the IEEE/CVF international conference on computer vision. pp. 3828--3838 (2019)

\bibitem{gordon2019depth}
Gordon, A., Li, H., Jonschkowski, R., Angelova, A.: Depth from videos in the wild: Unsupervised monocular depth learning from unknown cameras. In: Proceedings of the IEEE/CVF International Conference on Computer Vision. pp. 8977--8986 (2019)

\bibitem{grasa2013visual}
Grasa, O.G., Bernal, E., Casado, S., Gil, I., Montiel, J.: Visual slam for handheld monocular endoscope. IEEE transactions on medical imaging  \textbf{33}(1),  135--146 (2013)

\bibitem{liu2024dora}
Liu, S.Y., Wang, C.Y., Yin, H., Molchanov, P., Wang, Y.C.F., Cheng, K.T., Chen, M.H.: Dora: Weight-decomposed low-rank adaptation. In: Forty-first International Conference on Machine Learning (2024)

\bibitem{mountney2011}
Mountney, P.e.a.: Endoscopic camera calibration and its application in laparoscopic surgery. In: MICCAI 2011, LNCS. vol.~6891, pp. 473--480 (2011)

\bibitem{oquab2023dinov2}
Oquab, M., Darcet, T., Moutakanni, T., Vo, H., Szafraniec, M., Khalidov, V., Fernandez, P., Haziza, D., Massa, F., El-Nouby, A., et~al.: Dinov2: Learning robust visual features without supervision. arXiv preprint arXiv:2304.07193  (2023)

\bibitem{ozyoruk2021endoslam}
Ozyoruk, K.B., Gokceler, G.I., Bobrow, T.L., Coskun, G., Incetan, K., Almalioglu, Y., Mahmood, F., Curto, E., Perdigoto, L., Oliveira, M., et~al.: Endoslam dataset and an unsupervised monocular visual odometry and depth estimation approach for endoscopic videos. Medical image analysis  \textbf{71},  102058 (2021)

\bibitem{paruchuri2024leveraging}
Paruchuri, A., Ehrenstein, S., Wang, S., Fried, I., Pizer, S.M., Niethammer, M., Sengupta, R.: Leveraging near-field lighting for monocular depth estimation from endoscopy videos. In: European Conference on Computer Vision. pp. 473--491. Springer (2024)

\bibitem{ranftl2021vision}
Ranftl, R., Bochkovskiy, A., Koltun, V.: Vision transformers for dense prediction. In: Proceedings of the IEEE/CVF international conference on computer vision. pp. 12179--12188 (2021)

\bibitem{rattanalappaiboon2015fuzzy}
Rattanalappaiboon, S., Bhongmakapat, T., Ritthipravat, P.: Fuzzy zoning for feature matching technique in 3d reconstruction of nasal endoscopic images. Computers in Biology and Medicine  \textbf{67},  83--94 (2015)

\bibitem{recasens2021endo}
Recasens, D., Lamarca, J., F{\'a}cil, J.M., Montiel, J., Civera, J.: Endo-depth-and-motion: Reconstruction and tracking in endoscopic videos using depth networks and photometric constraints. IEEE Robotics and Automation Letters  \textbf{6}(4),  7225--7232 (2021)

\bibitem{shao2022self}
Shao, S., Pei, Z., Chen, W., Zhu, W., Wu, X., Sun, D., Zhang, B.: Self-supervised monocular depth and ego-motion estimation in endoscopy: Appearance flow to the rescue. Medical image analysis  \textbf{77},  102338 (2022)

\bibitem{solano2025multi}
Solano, P.E.C., Bulpitt, A., Subramanian, V., Ali, S.: Multi-task learning with cross-task consistency for improved depth estimation in colonoscopy. Medical Image Analysis  \textbf{99},  103379 (2025)

\bibitem{spencer2020defeat}
Spencer, J., Bowden, R., Hadfield, S.: Defeat-net: General monocular depth via simultaneous unsupervised representation learning. In: Proceedings of the IEEE/CVF Conference on Computer Vision and Pattern Recognition. pp. 14402--14413 (2020)

\bibitem{sun2023sc}
Sun, L., Bian, J.W., Zhan, H., Yin, W., Reid, I., Shen, C.: Sc-depthv3: Robust self-supervised monocular depth estimation for dynamic scenes. IEEE Transactions on Pattern Analysis and Machine Intelligence  (2023)

\bibitem{yang2024depth}
Yang, L., Kang, B., Huang, Z., Zhao, Z., Xu, X., Feng, J., Zhao, H.: Depth anything v2. Advances in Neural Information Processing Systems  \textbf{37},  21875--21911 (2024)

\bibitem{yang2024self}
Yang, Z., Pan, J., Dai, J., Sun, Z., Xiao, Y.: Self-supervised lightweight depth estimation in endoscopy combining cnn and transformer. IEEE Transactions on Medical Imaging  (2024)

\bibitem{zeinoddin2025endo}
Zeinoddin, M.S., Islam, M., Tandogdu, Z., Shaw, G., Clarkson, M.J., Mazomenos, E., Stoyanov, D.: Endo-fast3r: Endoscopic foundation model adaptation for structure from motion. arXiv preprint arXiv:2503.07204  (2025)

\bibitem{zeinoddin2024dares}
Zeinoddin, M.S., Lena, C., Qu, J., Carlini, L., Magro, M., Kim, S., De~Momi, E., Bano, S., Grech-Sollars, M., Mazomenos, E., et~al.: Dares: Depth anything in robotic endoscopic surgery with self-supervised vector-lora of the foundation model. arXiv preprint arXiv:2408.17433  (2024)

\bibitem{zhang2020real}
Zhang, P., Luo, H., Zhu, W., Yang, J., Zeng, N., Fan, Y., Wen, S., Xiang, N., Jia, F., Fang, C.: Real-time navigation for laparoscopic hepatectomy using image fusion of preoperative 3d surgical plan and intraoperative indocyanine green fluorescence imaging. Surgical endoscopy  \textbf{34},  3449--3459 (2020)

\end{thebibliography}
\bibliographystyle{splncs04}
\end{document}